\documentclass{article}

\usepackage{graphicx}
\usepackage{dblfnote}

\usepackage{float}
\usepackage{booktabs}
\usepackage{multirow}
\usepackage{booktabs, multirow} % for borders and merged ranges
\usepackage{soul}% for underlines
\usepackage[table]{xcolor} % for cell colors
\usepackage{changepage,threeparttable} % for wide tables
\usepackage[pagebackref=true,breaklinks=true,letterpaper=true,colorlinks,bookmarks=false]{hyperref}

    \usepackage[preprint]{neurips_data_2023}
\usepackage[utf8]{inputenc} % allow utf-8 input
\usepackage[T1]{fontenc}    % use 8-bit T1 fonts
\usepackage{hyperref}       % hyperlinks
\usepackage{url}            % simple URL typesetting
\usepackage{booktabs}       % professional-quality tables
\usepackage{amsfonts}       % blackboard math symbols
\usepackage{nicefrac}       % compact symbols for 1/2, etc.
\usepackage{microtype}      % microtypography
\usepackage{xcolor}         % colors

\makeatletter
\newcommand{\myfnsymbol}[1]{%
  \expandafter\@myfnsymbol\csname c@#1\endcsname
}
\newcommand{\@myfnsymbol}[1]{%
  \ifcase #1
  \or 1% 1
  \or 2% 2
  \or 3% 1
  \or 4% 2
  \or 5% 1
  \or 6% 2
  \or 7% 1
  \or 8% 2
  \or 9% 2
  \or \TextOrMath{\textasteriskcentered}{*}% 3
  \fi
}
\newcommand{\harvard}{\@myfnsymbol{1}}
\newcommand{\princeton}{\@myfnsymbol{2}}
\newcommand{\stanford}{\@myfnsymbol{3}}
\newcommand{\mgh}{\@myfnsymbol{4}}
\newcommand{\sbh}{\@myfnsymbol{5}}
\newcommand{\ut}{\@myfnsymbol{6}}
\newcommand{\uta}{\@myfnsymbol{7}}
\newcommand{\nyu}{\@myfnsymbol{8}}
\newcommand{\ucsf}{\@myfnsymbol{9}}
\newcommand{\equalcontributor}{\@myfnsymbol{10}}
\makeatother

\title{BenchMD: A Benchmark for Unified Learning
on Medical Images and Sensors}

\author{%
Kathryn Wantlin\textsuperscript{\harvard, \princeton},
\,
Chenwei Wu\textsuperscript{\harvard},
\,
Shih-Cheng Huang\textsuperscript{\stanford},
\,
Oishi Banerjee\textsuperscript{\harvard},
\,
Farah Dadabhoy\textsuperscript{\mgh},\\
\And
Veeral Vipin Mehta\textsuperscript{\sbh},
\,
Ryan Wonhee Han\textsuperscript{\stanford},
\,
Fang Cao\textsuperscript{\stanford},
\,
Raja R. Narayan\textsuperscript{\mgh},
\,
Errol Colak\textsuperscript{\ut},\\
\And
Adewole Adamson\textsuperscript{\uta},
\,
Laura Heacock\textsuperscript{\nyu},
\,
Geoffrey H. Tison\textsuperscript{\ucsf},
\,
Alex Tamkin\textsuperscript{\stanford,\equalcontributor},
\,
Pranav Rajpurkar\textsuperscript{\harvard, \equalcontributor}\\
}

\begin{document}

% Thanks notes for title uses \myfnsymbol
\renewcommand{\thefootnote}{\myfnsymbol{footnote}}
\maketitle

\footnotetext[1]{Harvard University}%
\footnotetext[2]{Princeton University}%
\footnotetext[3]{Stanford University}%
\footnotetext[4]{Massachusetts General Hospital}%
\footnotetext[5]{Stony Brook University Hospital}%
\footnotetext[6]{University of Toronto}%
\footnotetext[7]{University of Texas at Austin}%
\footnotetext[8]{NYU Langone Health}%
\footnotetext[9]{University of California, San Francisco}%
\footnotetext[10]{Equal senior authorship.}%

\setcounter{footnote}{0}% Restart footnote counter
% Footnotes for rest of document uses \fnsymbol (or whatever you choose)
\renewcommand{\thefootnote}{\fnsymbol{footnote}}

\maketitle

\begin{abstract}
Medical data poses a daunting challenge for AI algorithms: it exists in many different modalities, experiences frequent distribution shifts, and suffers from a scarcity of examples and labels. 
Recent advances, including transformers and self-supervised learning, promise a more universal approach that can be applied flexibly across these diverse conditions.
To measure and drive progress in this direction, we present BenchMD: a benchmark that tests how well unified, modality-agnostic methods, including architectures and training techniques (e.g. self-supervised learning, ImageNet pretraining),perform on a diverse array of clinically-relevant medical tasks. 
BenchMD combines 19 publicly available datasets for 7 medical modalities, including 1D sensor data, 2D images, and 3D volumetric scans. 
Our benchmark reflects real-world data constraints by evaluating methods across a range of dataset sizes, including challenging few-shot settings that incentivize the use of pretraining. 
Finally, we evaluate performance on out-of-distribution data collected at different hospitals than the training data, representing naturally-occurring distribution shifts that frequently degrade the performance of medical AI models. 
Our baseline results demonstrate that no unified learning technique achieves strong performance across all modalities, leaving ample room for improvement on the benchmark. Code is released at: {\small\url{https://github.com/rajpurkarlab/BenchMD}}. 
\end{abstract}

\section{Introduction}
\label{sec:intro}

Recent advances in transformers and self-supervised learning (SSL) have enabled state-of-the-art performance across many modalities, including text, images and videos \cite{Ericsson2022-zc, OpenAI2023-ce}. A core feature of these methods is their remarkable versatility: they lessen the need for labeled data, and can be applied flexibly across modalities, reducing the need to develop custom methods for each application area \cite{Tamkin2021-jw, NEURIPS2022_fa73aca7}. Measuring this progress requires benchmarks with \textit{breadth}, to capture the diversity of applications and modalities, as well as \textit{depth}, to ensure external validity by involving experts in the benchmark-formulation process \cite{Raji2021-hy, Bowman2021-ru, Ott2022-bc}. These requirements are especially salient in the medical domain, where existing benchmarks have been criticized for addressing synthetic tasks with low clinical relevance \cite{Blagec2022-do}.

\begin{figure*}
    \centering
    \includegraphics[width=0.8\textwidth]{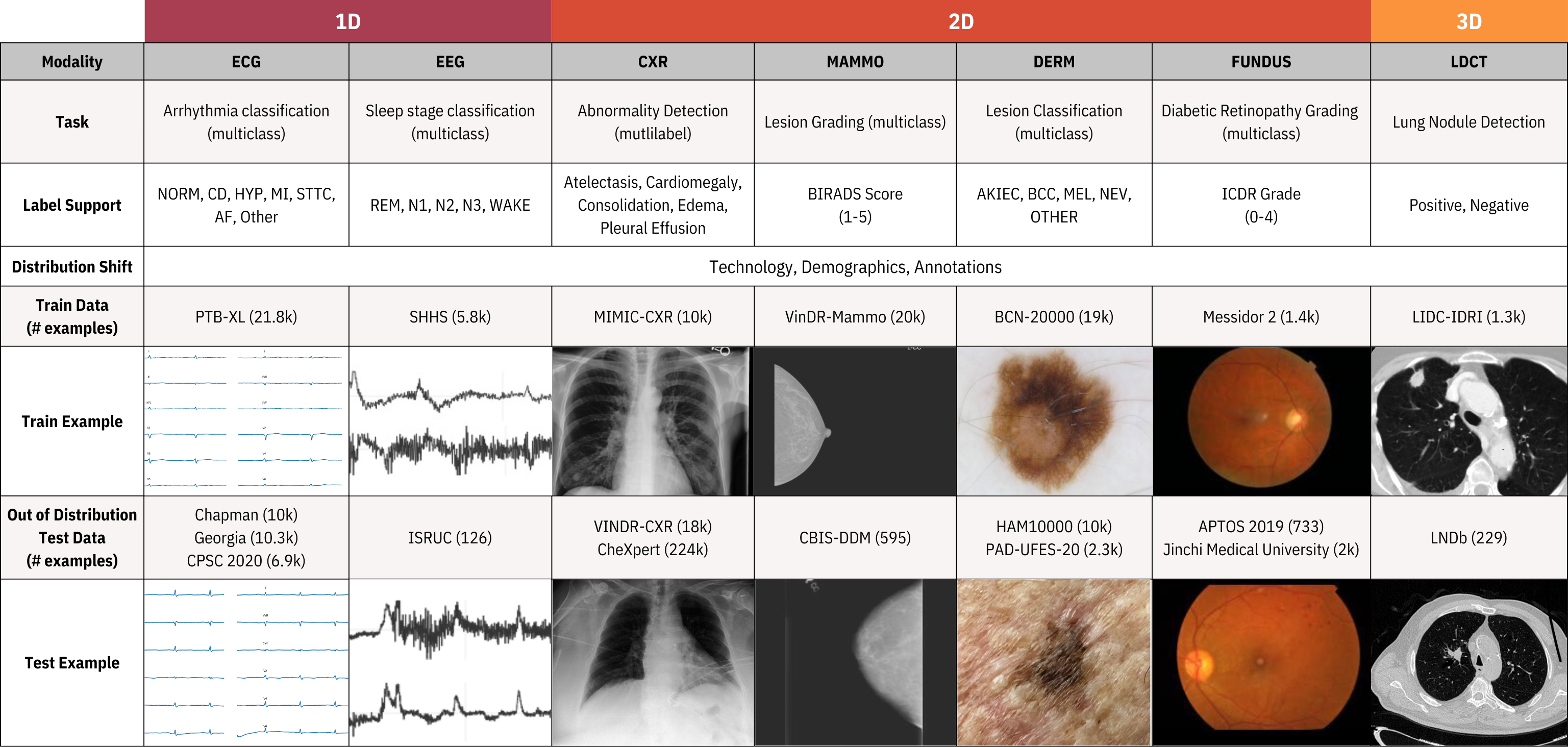}
    \caption{The BenchMD benchmark consists of 19 real-world medical datasets across 7 medical modalities. Successful methods will achieve high performance when evaluated on out-of-distribution data.
    }
    \label{fig:data}
\end{figure*}

To address this gap, we propose BenchMD, a new benchmark for unified learning across modalities that is grounded in real-world medical interpretation tasks and distribution shifts. BenchMD evaluates \textbf{unified architectures and training techniques} (e.g. SSL, ImageNet pretraining), on 19 datasets for 7 medical modalities. The wide variety of modalities reflects the heterogeneity of medical image and sensor data, which can be produced by dozens of different technologies \cite{Wolbarst2013-wx}. Specifically, we evaluate methods using 1D data from electrocardiogram (ECG) and electroencephalogram (EEG) sensors, 2D image data from chest X-rays (CXR), mammograms, dermoscopic images, and fundus images,  and 3D volumetric data from low-dose computed tomography (LDCT) scans (see Figure \ref{fig:data}). Current methods are often specialized for these different modalities; for example, contrastive learning techniques typically require modality-specific data augmentations \cite{Krishnan2022-fj}. Researchers for each medical application are similarly focused on designing domain-specific methods, using trial and error to establish which components (different architectures, self-supervision algorithms, etc.) will be helpful for their problem. In contrast, we encourage the development of \textbf{flexible, unified methods that can be applied out-of-the-box} without customization. Succeeding on arbitrary data requires unified architectures, and there is emerging evidence of promising candidates here \cite{Tamkin2021-jw,NEURIPS2022_fa73aca7,Koh2021-nv}. Our benchmark serves to accelerate progress in this area by enabling researchers to build domain-specific models from stronger baseline components and encouraging valuable collaboration on methods that are broadly useful across medical domains and modalities \cite{Parekh2021-rr}. Key to our unified approach, BenchMD constructs \textbf{standardized, clinically-impactful tasks} for evaluation in each modality, each validated by experts to have high  medical relevance given available label information. 

We construct our benchmark to enable advances on two additional fronts. First, label shortages have historically posed a serious obstacle to model development, so our benchmark tests performance under severe \textbf{data scarcity}, incentivizing the use of SSL techniques that exploit unlabeled data. In order to explore the label efficiency of different methods, we assess performance across multiple settings with different amounts of labeled data. Second, we investigate how models perform under \textbf{naturally-occurring distribution shifts}, such as when they are trained on data from one hospital and deployed in another. To this end, we train models on one in-distribution (ID) source dataset and then test zero-shot transfer performance on out-of-distribution (OOD) data from unseen target datasets collected at different hospitals. Thus, our definition of a successful ``unified'' method includes a universally superior training objective and architecture that is applicable across a range of different modalities and, in addition, generalizes well across distribution shifts within modalities. 

We present BenchMD as an easy-to-use benchmark for assessing performance widely across medical modalities and distributions. To make using BenchMD \textbf{simple}, we standardize preprocessing steps and validation metrics (details in the Appendix), so users simply need to plug in new architectures and training tasks. Additionally, we use only \textbf{publicly available datasets}, allowing users to easily access BenchMD and replicate results. Using our benchmark, we provide initial baselines that demonstrate significant variations in performance, with no technique achieving strong results across all modalities and ImageNet-pretrained baselines exceeding the performance of existing domain-agnostic SSL methods in some modalities. These results motivate the necessity of further research in universal, generalizable methods for medical AI, and we discuss possible directions for future work which are all easily facilitated by the publicly available code and data constructed for BenchMD. We expect our work will accelerate the development of versatile methods for medicine and provide a valuable tool for measuring advances in universal methods.

\begin{figure*}
    \centering
    \includegraphics[width=0.9\textwidth]{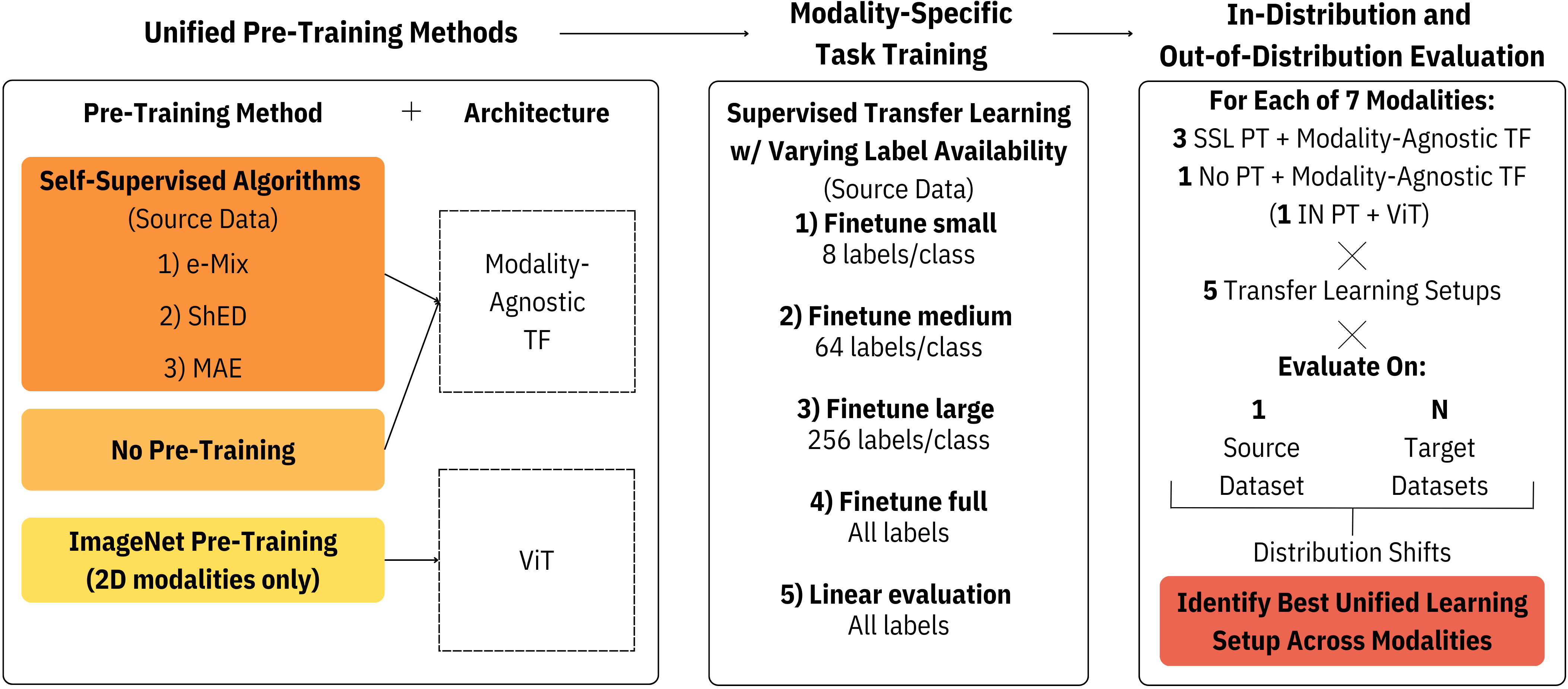}
    \caption{Models for each modality are first trained on a source dataset, using unified methods across modalities. They are then evaluated on out-of-distribution data from one or more target datasets.}
    \label{fig:methods}
\end{figure*}

\section{Related Work}
\label{sec:related}
\paragraph{Unified Techniques Across Modalities:} Recent advances in deep learning have produced methods that enable high performance and can be flexibly applied across modalities. Self-supervised learning (SSL) techniques such as masked data modeling \cite{Girdhar2022-oh} and contrastive learning \cite{Lee2020-st, tamkin2020viewmaker, Tamkin2021-jw} can be used to learn from unlabeled datasets across many modalities. Architectures are also increasingly able to take in different modalities as input, producing models that can interchangeably process 2D images and 3D videos \cite{jaegle2021perceiver, Likhosherstov2021-zn,NEURIPS2022_960a172b}. Benchmarks like ours can rigorously evaluate these new methods, assessing their performance on real-world tasks across multiple modalities.

\paragraph{Unified Medical AI:} There have been limited efforts to unify architectures and training techniques across medical image modalities in particular \cite{Zhou2021-lq,Zhai2019-pu,Ghesu2022-fa}. For example, Zhou et al. recently found that self-supervised MAE pretraining on medical images offered better performance than ImageNet pretraining when interpreting chest X-rays, MRI scans, and CT scans \cite{Zhou2022-oa}. Similarly, Azizi et al. found that a training procedure combining supervised learning on natural images with SSL pretraining on medical images offered high performance on 6 2D medical image modalities \cite{Azizi2022-jm}. BenchMD tracks progress in this area and includes an unprecedented range of medical modalities, addressing a range of 1D, 2D, and 3D medical images and sensors.

\paragraph{Existing Benchmarks for Multiple Modalities:} Our work extends the line of thinking in the DABS benchmarks, which evaluate the performance of modality-agnostic SSL techniques across modalities ranging from text and genomics to X-rays and wearable sensor data \cite{Tamkin2021-jw,NEURIPS2022_fa73aca7}. We also take inspiration from the WILDS benchmarks, which evaluate performance on out-of-distribution data within several modalities \cite{Koh2021-nv,Sagawa2021-ol,Yao2022-xh}. BenchMD combines the strengths of both approaches, creating a benchmark that is rooted in real-world modalities with direct clinical applicability: evaluating whether techniques generalize well across modalities as well as to new distributions within each modality. Furthermore, while some of DABS’s training datasets are unlabeled, we provide labeled source data to facilitate comparison against non-SSL techniques such as supervised learning. Unlike both DABS and WILDS, BenchMD tests model performance across settings with few-shot learning, exploring how label availability affects performance. Our work also differs from both DABS and WILDS because we focus on real-world medical tasks and cover a broader range of medical modalities, including 3D volumetric scans.

\section{Modalities and Datasets}
\label{sec:mod}

We have curated a list of high-impact modalities and selected source and target datasets for evaluating out-of-distribution (OOD) performance. Each modality we present in this benchmark is used to test for prevalent diseases and significantly contributes to clinician workloads in current practice (see \cite{Zhu2020-lq, chiao2017trends, NAP11617, Hoorens2019-jy, irvin2019chexpert, Monticciolo2021-ua, Abramoff2016-pc, Schreuder2021-mk}). For each modality, we select a highly-cited, large dataset as the source dataset, whose training and validation splits we use for pretraining and in-distribution evaluation, respectively. We also choose labeled target datasets, whose validation sets we use to test performance on OOD data. The datasets, in addition to all being well-known and publicly accessible, were selected to cover a range of important distribution shifts for each modality, including varying demographics, collection technology, and annotation details. Per modality, we specify a task that is both clinically relevant and unified across source and target datasets. (See the supplementary appendix.)

\subsection{Electrocardiograms}
12-lead electrocardiogram (ECG) measures the three-dimensional electrical activity of the heart over time using electrodes placed on the skin. Classifying cardiovascular abnormalities from ECGs is challenging because there are 12 1D channels, each corresponding to a different spatial axis, and because diagnosis requires distinguishing irregular cardiovascular signals from noisy data. We perform a single-label, 7-class classification task, with inputs consisting of 5 second recordings of 12-channel ECG signals with a sampling rate of 500Hz. Our set of 7 labels is unified across datasets and derived at the discretion of medical experts: Normal, Conduction Disturbance, Hypertrophy, Myocardial Infarction, Ischemic ST-T Changes, Atrial fibrillation/atrial flutter, and Other. We consider four publicly available 12-lead ECG-waveform datasets: PTB-XL (source) \cite{wagner2020ptb, Wagner2022-hc, Goldberger2000-ex}, Chapman-Shaoxing (target)\cite{zheng202012}, Georgia 12-Lead ECG Challenge (target)\cite{goldberger2000physiobank}, and China Physiological Signal Challenge (CPSC, target)\cite{cai2020open}.
We see the following distribution shifts: \textit{Demographics:} PTB-XL's data was collected between 1989 and 1996, Chapman-Shaoxing's in 2020, CPSC's in 2018, and Georgia's in 2020. The Chapman-Shaoxing and CPSC datasets’ patients are based in China, the Georgia datasets’ in the southeastern United States, and the PTB-XL datasets’ in Germany. 
\textit{Collection Technology:} The PTB-XL dataset used devices provided by Schiller AG, while the Chapman-Shaoxing dataset was collected using devices from Zhejiang Cachet Jetboom Medical Devices. 
\textit{Annotation Details:} Although we group abnormalities into 7 categories that are consistent across datasets, different datasets provide varying levels of additional granularity in their labels, with different label distributions across datasets. 

\subsection{Electroencephalograms}
Electroencephalograms (EEG) measure multi-channel 1D signals of electrical activity in the brain and are used to diagnose sleep and seizure disorders\cite{Bandyopadhyay2022-tf}. Noise and intraclass variability in sampling rates, signal quality, the number of leads used, and the length of the captured rhythm makes distinguishing sleep stages difficult in EEGs. We perform a single-label sleep stage classification task on 2 traditional central derivations channels (C3 and C4), 125 Hz, 30 second EEG signal inputs. We use the American Academy of Sleep Medicine's standard 5 labels: Wake, Rapid Eye Movement, Non-REM Stage 1, Non- REM stage 2, and Non-REM stage 3\cite{Duce2014-gl}. We consider two publicly available datasets: the Sleep Heart Health Study (SHHS) dataset (source) \cite{quan1997sleep} and the ISRUC-Sleep dataset (target) \cite{khalighi2016isruc}. 
We see the following distribution shifts: \textit{Demographics:} The SHHS dataset was collected between 1995-1998, while the ISRUC dataset was collected between 2009–2013. The SHHS dataset includes 5,804 adults aged 40 and older, while the ISRUC dataset was collected from subjects whose ages range from 20 years old to 85 years old, with an average age of 51. Moreover, the ISRUC dataset was collected from a hospital in Coimbra, Portugal, while SHHS was collected by the National Heart Lung \& Blood Institute in the US.
\textit{Collection Technology:}
The SHHS dataset was collected at a sampling rate of 125 Hz while ISRUC was collected at 150 Hz. SHHS was also collected from studies conducted in patient homes, while ISRUC was collected in a hospital setting.

\subsection{Chest X-Rays}

Chest X-rays are 2D grayscale projection radiographs of a patient's heart, lungs, blood vessels, airways, chest bones, and spine and are crucial for the diagnosis of cardiovascular diseases such as atelectasis and edema. Chest X-ray classification is uniquely challenging compared to natural image classification since radiographs are grayscale and always have similar frontal or lateral spatial structures, with relevant abnormalities only occurring in a small region of the image\cite{Ke2021-qv}. We perform a single-label classification task on 2D grayscale chest x-rays using 5 prevalent labels: Atelectasis, Cardiomegaly, Consolidation, Edema, and Pleural Effusion. We utilize three publicly available datasets: MIMIC-CXR (source)\cite{johnson2019mimic, Johnson2019-nx, goldberger2000physiobank}, CheXpert (target)\cite{irvin2019chexpert}, and VinDr-CXR (target)\cite{nguyen2022vindr}. 
We see the following distribution shifts: \textit{Demographics:} MIMIC-CXR's images were collected between 2011 and 2016, CheXpert's between 2002 and 2017, and VinDr-CXR's between 2018 and 2020. CheXpert's data was collected by Stanford University School of Medicine, MIMIC-CXR's from the Beth Israel Deaconess Medical Center in Boston, and VinDr-CXR's from the Hanoi Medical University Hospital and Hospital 108 in Vietnam.
\textit{Annotation Details:} MIMIC-CXR and CheXpert use automated natural language labelers, while VinDr-CXR uses radiologist-generated annotations.

\subsection{Mammograms}
Mammograms consist of 2D grayscale images of the cranio-caudal (CC) view and the mediolateral-oblique (MLO) view of the left and right breast of a patient (4 images possible per patient), and are the main imaging tool for the screening and diagnosis of breast cancer\cite{Boumaraf2020-zj}. Mammograms are high-resolution images with millions of pixels, and while the breast views are highly standardized, disease classification depends on abnormalities in small regions of interest, making diagnosis challenging for AI models. We perform a single-label task of predicting the Breast Imaging Reporting and Data System (BI-RADS) assessment category (from 1 to 5) for each breast image. We consider two datasets: VinDr-Mammo (source) \cite{nguyen2022vinmammo} and CBIS-DDSM (target)\cite{lee2017curatedcbis, Clark2013-vf, Smith_undated-gg}.
We see the following distribution shifts: \textit{Demographics:} VinDr-Mammo was compiled from a pool of mammography examinations taken between 2018 and 2020, while CBIS-DDSM was compiled from exams conducted between 1988 and 1999. VinDr-Mammo exams were collected by hospitals in Vietnam, while CBIS-DDSM’s were collected by United States hospitals.
\textit{Collection Technology:} VinDr-Mammo contains full-field digital mammogram images while CBIS-DDSM contains scanned film mammogram images. Several different scanners from multiple manufacturers were used to collect the CBIS-DDSM mammograms. 
\textit{Annotation Details:} VinDr-Mammo contains some lesionless images, which still have a breast-level BI-RADS score. The CBIS-DDSM dataset, however, exclusively contains images with lesions, and does not annotate breast-level BI-RADS scores. To test our model on this target dataset, for each breast we use the maximum of lesion-level BI-RADS scores as the breast-level BI-RADS score. 

\subsection{Dermoscopic Images}
Dermoscopy produces 2D RGB images showing subsurface skin structures in the epidermis, at the dermoepidermal junction, and in the papillary dermis, and is used to assess cancer in skin lesions. Performing tasks on dermoscopic images is complicated by intraclass variability in lesion texture, scale, and color due to presence of different skin colors, hair, veins, and irregular lesion borders \cite{Hoorens2019-jy}. We perform a single-label classification of 2D RGB dermoscopy images across 5 unified labels extracted by clinicians: ``AKIEC" (includes actinic keratoses, intraepithelial carcinoma, and squamous cell carcinoma as all of these are with the continuum of squamous cell carcinoma), ``BCC" (basal cell carcinoma), ``MEL" (melanoma), ``NEV" (nevus), and ``Other diseases" ( dermatofibroma, etc). We utilize three publicly available datasets: BCN 20000 (source) \cite{combalia2019bcn20000}, HAM 10000 (target)\cite{tschandl2018ham10000}, PAD-UFES-20 Smartphone image-set (target) \cite{pacheco2020pad}.
We see the following distribution shifts: \textit{Demographics:} BCN20000's images were collected from 2010 to 2016, PAD-UEFS-20's from 2020, and HAM10000's from the past 20 years. PAD-UEFS-20's images were collected by hospitals in Brazil, HAM10000's  in Austria and Australia, and BCN20000’s in Spain. 
\textit{Collection Technology:} BCN20000 and HAM10000 images were collected using dermatoscopes, while PAD-UFES-20 images were collected by smartphone cameras. 
\textit{Annotation Details:} Although we grouped the abnormality annotations across datasets into 5 general categories, the granularity within each label varies depending on the dataset. For example, the ``Other diseases" category for HAM10000 includes benign keratosis-like lesions while BCN20000's doesn’t.

\subsection{Fundus Images}
Eye fundus images are 2D RGB images showing the interior surface of a single eye, including the retina, fovea, optic disc, macula, and posterior pole, and are crucial for the diagnosis of diabetic retinopathy (DR). The detection of DR is complicated by spurious correlations with other undetected conditions such as diabetic macular edema\cite{Abramoff2016-pc}. For each 2D RGB fundus image, we perform the single-label task of predicting the severity of diabetic retinopathy (DR) in an image of each eye. We use the International Clinic Diabetic Retinopathy (ICDR) classification scale, which classifies DR on a five-stage severity scale from 0-4\cite{Ramchandre2020-mh}. We consider three datasets: Messidor-2 (source)\cite{decenciere2014feedbackmessi1,abramoff2013automatedmessi2}, APTOS 2019 (target)\cite{noauthor_undated-nv, noauthor_undated-pe}, and the Jinchi Medical University dataset (target)\cite{Takahashi2017-nn}. 
We see the following distribution shifts: \textit{Demographics:} Messidor-2 images were collected from 2004 to 2010, while Jinchi Medical University images were collected between May 2011 and June 2015. The total collection period for APTOS 2019 is unknown. The Messidor2 data was collected from French institutions, APTOS 2019 data from the Aravind Eye Care System in India, and the Jinchi Medical University data from Japan. The Messidor-2 and Jinchi Medical University datasets consist of high quality retinal images, while APTOS 2019 exhibits more variation in data quality, including images with artifacts.
\textit{Collection Technology:} Messidor-2 training images were taken with a Topcon TRC NW6 non-mydriatic camera. The Jinchi Medical University dataset also uses a non-mydriatic camera, but a different model (AFC-230). The APTOS dataset contains images taken from both mydriatic and non-mydriatic cameras, with the full range of camera models unknown. The Jinchi Medical University dataset was collected in a single-site, exploratory study performed in an institutional setting, whereas the other datasets contain images taken in clinical settings for diagnostic purposes.
\textit{Annotation Details:} Jinchi Medical University, unlike Messidor-2 and APTOS, consolidates the similar ICDR classes 1 and 2 into a single superclass, termed a modified Davis grading.

\subsection{Low Dose Computed Tomography Scans}
Low dose computed tomography (LDCT) is a procedure that uses an x-ray machine linked with a computer to create 3D images of a patient's tissues and organs. LDCT is typically used to detect early-stage nodules of lung cancer in high-risk patients. The LDCT nodule classification task is challenging since LDCT scans are 3D images originally recorded in single-channel Hounsfield units with varying numbers of slices between patients. In addition, while scans have a large field of view with hundreds of slices, nodules only occupy a small volume of the scan, especially in early cancer stages\cite{immonen2021use}. Inputs are partitioned by sliding windows, representing 24 CT slices in single channel Hounsfield units. We perform two binary classification tasks, determining 1) whether a small nodule (diameter $\leq$3mm) exists in the current CT scan window and 2) whether a large nodule (diameter $\leq$3mm) exists in the current CT scan window. A sliding window is labeled positive for a nodule of either type if it contains more than 4 consecutive slices with positive labels. Final determination, at the volume level, for both small and large nodule(s) presence is done by aggregating prediction probabilities from all windows. We utilize two public datasets: LIDC-IDRI (source)\cite{armato2011lunglidc} and LNDb (target)\cite{pedrosa2019lndb}.
We see the following distribution shifts: \textit{Demographics:} LIDC scans were collected in 2010, while LNDb scans were collected from 2016-2018. LIDC was collected from academic centers and medical imaging companies in the United States, while LNDb was collected at the Centro Hospitalar e Universitário de São João (CHUSJ) in Porto, Portugal.
\textit{Collection Technology:} LIDC dataset collection involved a variety of scanner manufacturers and models, while the LNDb dataset was primarily collected by Siemens scanners. LIDC's data was collected using a mean tube current of 222.1mA, while LNDb use a mean tube current of 161.9mA. The LIDC dataset includes slice thicknesses ranging from 0.6mm to 5mm, while the LNDb dataset has excluded CT scans where intravenous contrast had been used and those with a slice thickness greater than 1mm. 

\begin{figure*}
    \centering
    \includegraphics[width=0.8\textwidth]{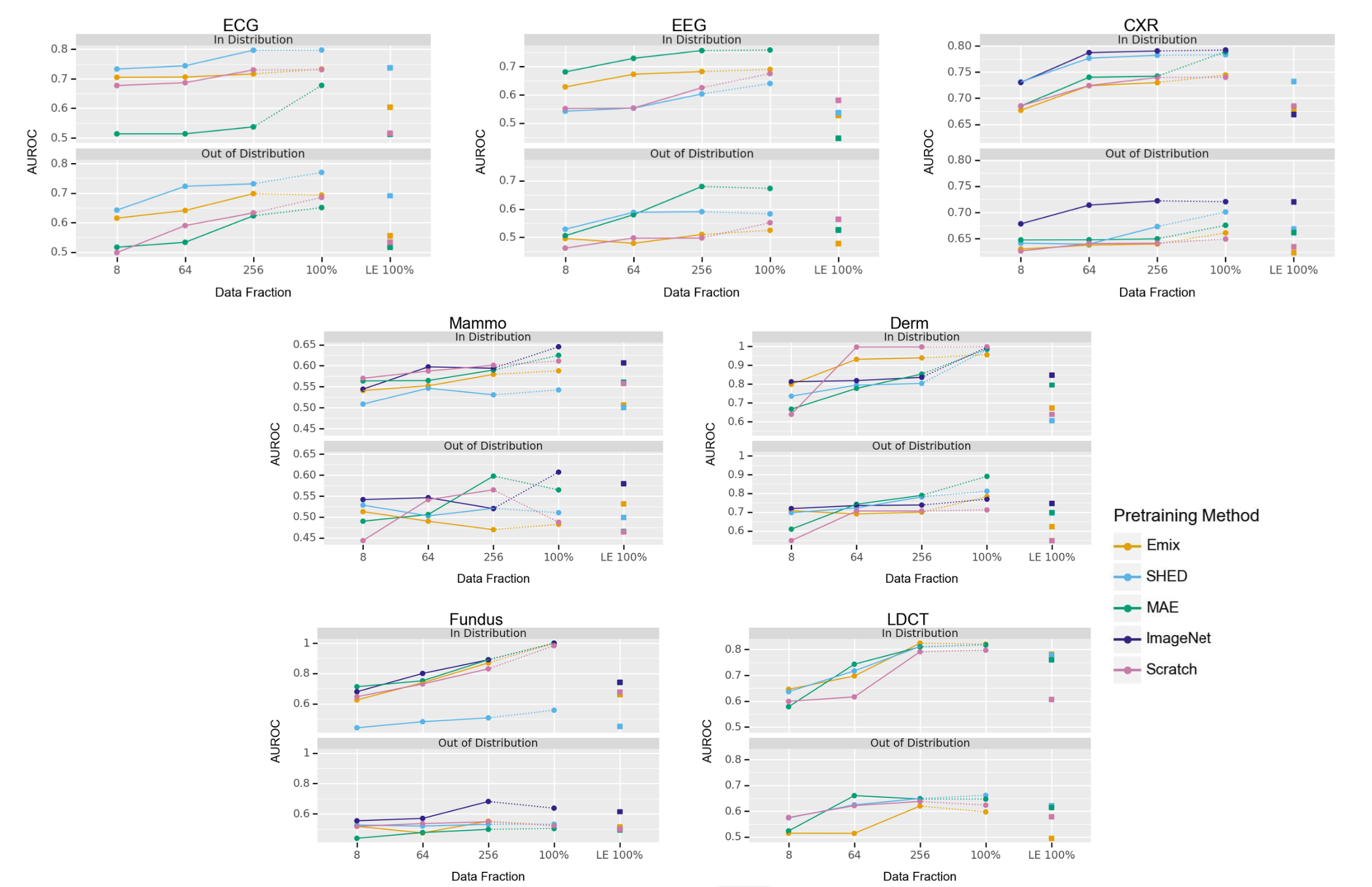}
    \caption{The in-distribution and out-of-distribution performance of models across modalities. OOD performance is averaged across target dataset(s).}
    \label{fig:results1}
\end{figure*}

\section{Experiments}
\label{sec:setup}

We evaluate the performance of five baseline techniques: three SSL algorithms, ImageNet pretraining, and training from scratch. We then test performance on OOD target datasets using multiple transfer learning schemes.  

\subsection{Architecture}

Following \cite{Tamkin2021-jw}, we utilize a modality-agnostic transformer architecture across all experiments. We use separate 1D, 2D, and 3D embedding modules, which make minimal assumptions about the data and map all inputs to the same 256-dimensional embedding space, allowing users to mix inputs with different input dimensions. The encoder is based on a standard vision transformer architecture \cite{vaswani2017attention, dosovitskiy2020image}, and we choose 1D, 2D, or 3D patch sizes to keep the resulting sequence length similar across all datasets. Additional details on the architecture are available in Appendix B.

\subsection{Pretraining}

We evaluate the performance of three different SSL algorithms in this benchmark. The first two, Contrastive Embedding-Mixup \textbf{(e-Mix)} and Shuffled Embedding Prediction \textbf{(ShED)}, follow \cite{Tamkin2021-jw}. e-Mix is a contrastive objective that additively mixes a batch of original input embeddings, weighting them with different coefficients. It then trains an encoder to produce a vector for a mixed embedding that is close to the original inputs' embeddings in proportion to their mixing coefficients. ShED shuffles a fraction (0.85 in our experiments) of embeddings and trains the encoder with a classifier to predict which embeddings were perturbed. Following the training settings used in \cite{NEURIPS2022_fa73aca7}, we also use a third Masked Autoencoding \textbf{(MAE)} objective, which masks a given fraction (0.75 in our experiments) of input embeddings and trains models to reconstruct them \cite{He2021-jn}. We standardize the pretraining process, running it for 100k steps with the Adam optimizer, learning rate 1e-4, weight decay 1e-4, and momentum 0.9. Beyond SSL, we evaluate two other techniques. First, we consider a \textbf{scratch} baseline, where the model is not pretrained. In addition, for 2D modalities, we evaluate models \textbf{pretrained on ImageNet}.

\subsection{Transfer Learning and Out-of-Distribution Evaluation}

We train models for particular tasks through both linear evaluation and finetuning, using labeled data from our in-distribution source datasets. We then evaluate zero-shot performance on OOD target datasets.
For linear evaluation, we freeze the model backbone and train a linear classifier head for the modality task using 100\% of the source data labels. For finetuning, we run one set of experiments using 100\% of the source labels but also test performance while varying label availability. For single-label tasks, we run experiments using 8, 64, or 256 labels per class, which we refer to as small, medium, and large label fractions, respectively. If the source dataset contains a class for which we have insufficient labels, we simply use all available labels for that class. For multi-label tasks, we create small/medium/large label sets by iterating through each class label and sampling labeled examples until we have 8, 64, or 256 labels for that class or have exhausted the available examples for that class. During both linear evaluation and finetuning, we train for 100 epochs with the Adam optimizer, learning rate 1e-4, weight decay 1e-4, and momentum 0.9. We evaluate our models using AUROC score as the metric (taking an unweighted average of per-class scores for multi-class tasks).

We then evaluate zero-shot transfer performance on OOD target datasets. After every epoch of linear evaluation or finetuning, we check the current model checkpoint's performance on the source dataset's validation set in order to perform model selection. We identify the top-performing checkpoint that achieves the highest average AUROC across tasks on the source validation set and report this AUROC in Figure ~\ref{fig:results1} under ``In Distribution". Next, we perform zero-shot transfer with this top-performing checkpoint by directly evaluating it on the target dataset(s), without any further training. We report AUROC on OOD data for each modality, averaged across tasks and target datasets, in Figure ~\ref{fig:results1}.

\subsection{Results}
Figure ~\ref{fig:results1} shows the performance of different techniques on the ID validation set and on our OOD test data. 
% Figure ~\ref{fig:shifts} shows how well different techniques perform and generalize, compared to training from scratch. We calculate the difference in performance between our pretraining techniques and training from scratch and compare how distribution shifts affect this difference. 
Overall, we find that no one method strictly dominates the others.

\paragraph{Does any SSL technique offer high performance across modalities?} No. e-Mix typically offers middling performance on OOD data; it outperforms other techniques on only on a few scattered experiments across ECG data, dermoscopic images and retinal fundus images. ShED is more promising, with particularly strong performance on ECG data and LDCT scans. However, it fails to maintain consistent performance across other modalities such as fundus images and mammograms, where ImageNet pretraining outperforms it across nearly all settings. Similarly, MAE achieves strong results on many experiments, especially on EEG data and on dermoscopic images from the PAD-UFES-20 dataset, it also performs poorly elsewhere. Despite being a top performer on EEG data, MAE achieves poor AUROC scores on ECG data, particularly the Georgia 12-Lead ECG Challenge and Chapman-Shaoxing datasets. This discrepancy indicates the difficulty of developing a single, high-performing technique even across different 1D sensor modalities. We see similar inconsistencies across different 2D image modalities, and future work may explore whether other SSL techniques or architectures offer more consistent performance.

\paragraph{Does any other technique offer high performance across modalities?} No. We investigate the use of ImageNet pretraining on 2D modalities and find it performs well across several modalities, typically outperforming other techniques on CXRs, mammograms, and fundus images from the APTOS dataset. However, SSL methods sometimes outperform ImageNet pretraining, with a particularly large gap on OOD dermoscopic images. Additionally, while models trained from scratch are rarely top-performers in any experiment, they still remain competitive, frequently outperforming at least one other model. The fact that ImageNet pretraining and training from scratch can sometimes match SSL performance demonstrates the difficulty of using SSL techniques out-of-the-box, without customization for particular medical modalities. We further explored a two-stage approach on the 2D mammogram and fundus image modalities, pre-training models on ImageNet before performing SSL using the MAE objective with some hyperparameter tuning, and found that this approach can yield benefits over either IN or SSL alone. (See supplement appendix tables.) Furthermore, while we standardized training times by performing the same number of iterations across all modalities, future work may explore other ways to set these and other hyperparameters. 

\paragraph{Does label availability affect performance?} Yes. Across all techniques and modalities, OOD performance typically stays the same or improves when more labels are available, though we see rare exceptions. For instance, MAE performance on mammograms drops when finetuning on 100\% of the data, suggesting that the model may have overfit. Once again, future work may benefit from dynamically adjusting the finetuning process to prevent overfitting.

\paragraph{How does performance change across distributions?} Though we sometimes see promising generalization performance, there are also cases where performance drops on OOD datasets. On the 100\% fine-tuning settings for ECG data, EEG data and mammograms, the top-performing technique on the in-distribution validation set also achieves the best performance across all OOD datasets, suggesting that generalization is successful. However, we see other cases of performance degradation due to distribution shift. For example, training from scratch achieves near-perfect performance on in-distribution dermoscopic images when fine-tuned with 64 or more data points. However, these models generalize poorly to OOD dermoscopic data, where training from scratch is never the top-performing technique on any experiment. Similarly, e-Mix is the top performer on most in-distribution LDCT experiments, yet it performs worst on all experiments using OOD data. Future work may use regularization techniques to improve generalization performance.

\section{Limitations}

While BenchMD generally aims to treat all modalities the same, we follow DABS's example in providing different embedding modules for 1D, 2D, and 3D data \cite{Tamkin2021-jw}. Users can replace these modules with their own, and we are hopeful that future approaches will identify ways to unify even this step across modalities. Additionally, while SSL has demonstrated promise in the medical domain, our SSL baselines achieve modest performance and fail to provide consistent benefits over ImageNet pretraining on 2D modalities. This may be because prior objectives do not generalize well across the diverse domains we consider. The ImageNet training technique we present is also inherently limited, as it is only appropriate for 2D images; as pretraining on natural images appears to offer benefits, future work may extend this approach to 1D and 3D data, such as by incorporating natural video pretraining as well. Our benchmark also currently allows easy modification of architectures and dataset types used in the training pipeline, flexibly allowing for future training with joint modality datasets and domain-agnostic Perceivers or CNNs. Finally, while we endeavor to cover a diverse range of medical modalities and datasets, it is impossible to fully represent the breadth of data in the medical domain. To protect patient safety, medical AI models should undergo further validation, such as through site-specific testing, before being deployed.

\section{Conclusion}
\label{sec:ccl}

We present BenchMD, a benchmark for evaluating unified methods across medical image and sensor modalities. While our initial baselines show some potential, there are ample opportunities for future work to improve both versatility and performance. Methods that succeed on BenchMD may also be applicable to many other modalities and distributions and can have real-world impact on clinical practice. We hope BenchMD will help promote the development of high-performing, generalizable and label-efficient methods for universal learning.

\begin{ack}
This project was supported by AWS Promotional Credits and by Harvard Data Science Institute Competitive Research Award. AT is supported by an Open Phil AI Fellowship.
\end{ack}

\bibliographystyle{abbrvnat}
\bibliography{references}

\appendix
\newpage

\section{Additional Results}
See Table \ref{tab:res} and \ref{tab:res-inpt} for our full set of results.

%Please add the following packages if necessary:
%\usepackage{booktabs, multirow} % for borders and merged ranges
%\usepackage{soul}% for underlines
%\usepackage[table]{xcolor} % for cell colors
%\usepackage{changepage,threeparttable} % for wide tables
%If the table is too wide, replace \begin{table}[!htp]...\end{table} with
%\begin{adjustwidth}{-2.5 cm}{-2.5 cm}\centering\begin{threeparttable}[!htb]...\end{threeparttable}\end{adjustwidth}
\begin{table*}[!htp]\centering
\caption{AUROC achieved for each modality: electrocardiograms (ECG), electroencephalograms (EEG), chest X-rays (CXR), mammo-
grams (Mammo), dermoscopic images (Derm), fundus images (Fundus), and low-dose computed tomography (LDCT) scans. The source
and target datasets are labelled accordingly. We consider performance of each pretraining objective along the columns: no pretraining
(scratch), e-Mix, ShED, MAE, and ImageNet pretraining (IN). For each modality we report results for each transfer learning setup along
the rows: linear evaluation using 100% of the labeled source data (LE) and finetuning with small/medium/large/100% label sets (FT-S
/ FT-M / FT-L / FT, respectively). For each dataset and transfer learning setup, we bold the max AUROC achieved across the different
pre-training methods.}\label{tab:res}
\scriptsize
\begin{tabular}{lrrrrr|rrrrr}\toprule
&Scratch &e-Mix &ShED &MAE &IN &Scratch &e-Mix &ShED &MAE &IN \\
\cellcolor[HTML]{efefef}\textbf{ECG } &\multicolumn{5}{c|}{\cellcolor[HTML]{efefef}PTB-XL (source)} &\multicolumn{5}{c}{\cellcolor[HTML]{efefef}Chapman-Shaoxing (target)} \\
\cellcolor[HTML]{efefef}LE &\cellcolor[HTML]{efefef}0.516 &\cellcolor[HTML]{efefef}0.604 &\cellcolor[HTML]{efefef}\textbf{0.737} &\cellcolor[HTML]{efefef}0.513 &\cellcolor[HTML]{efefef}- &\cellcolor[HTML]{efefef}0.551 &\cellcolor[HTML]{efefef}0.569 &\cellcolor[HTML]{efefef}\textbf{0.694} &\cellcolor[HTML]{efefef}0.529 &\cellcolor[HTML]{efefef}- \\
\cellcolor[HTML]{efefef}FT-S &\cellcolor[HTML]{efefef}0.677 &\cellcolor[HTML]{efefef}0.705 &\cellcolor[HTML]{efefef}\textbf{0.733} &\cellcolor[HTML]{efefef}0.514 &\cellcolor[HTML]{efefef}- &\cellcolor[HTML]{efefef}0.504 &\cellcolor[HTML]{efefef}0.703 &\cellcolor[HTML]{efefef}\textbf{0.723} &\cellcolor[HTML]{efefef}0.530 &\cellcolor[HTML]{efefef}- \\
\cellcolor[HTML]{efefef}FT-M &\cellcolor[HTML]{efefef}0.687 &\cellcolor[HTML]{efefef}0.706 &\cellcolor[HTML]{efefef}\textbf{0.744} &\cellcolor[HTML]{efefef}0.514 &\cellcolor[HTML]{efefef}- &\cellcolor[HTML]{efefef}0.650 &\cellcolor[HTML]{efefef}0.665 &\cellcolor[HTML]{efefef}\textbf{0.758} &\cellcolor[HTML]{efefef}0.535 &\cellcolor[HTML]{efefef}- \\
\cellcolor[HTML]{efefef}FT-L &\cellcolor[HTML]{efefef}0.730 &\cellcolor[HTML]{efefef}0.716 &\cellcolor[HTML]{efefef}\textbf{0.797} &\cellcolor[HTML]{efefef}0.537 &\cellcolor[HTML]{efefef}- &\cellcolor[HTML]{efefef}0.672 &\cellcolor[HTML]{efefef}0.744 &\cellcolor[HTML]{efefef}\textbf{0.762} &\cellcolor[HTML]{efefef}0.655 &\cellcolor[HTML]{efefef}- \\
\cellcolor[HTML]{efefef}FT &\cellcolor[HTML]{efefef}0.731 &\cellcolor[HTML]{efefef}0.733 &\cellcolor[HTML]{efefef}\textbf{0.797} &\cellcolor[HTML]{efefef}0.678 &\cellcolor[HTML]{efefef}- &\cellcolor[HTML]{efefef}0.711 &\cellcolor[HTML]{efefef}0.752 &\cellcolor[HTML]{efefef}\textbf{0.806} &\cellcolor[HTML]{efefef}0.656 &\cellcolor[HTML]{efefef}- \\
\cellcolor[HTML]{efefef}\textbf{ECG } &\multicolumn{5}{c|}{\cellcolor[HTML]{efefef}Georgia (target)} &\multicolumn{5}{c}{\cellcolor[HTML]{efefef}CPSC (target)} \\
\cellcolor[HTML]{efefef}LE &\cellcolor[HTML]{efefef}0.527 &\cellcolor[HTML]{efefef}0.566 &\cellcolor[HTML]{efefef}\textbf{0.692} &\cellcolor[HTML]{efefef}0.492 &\cellcolor[HTML]{efefef}- &\cellcolor[HTML]{efefef}0.523 &\cellcolor[HTML]{efefef}0.532 &\cellcolor[HTML]{efefef}\textbf{0.687} &\cellcolor[HTML]{efefef}0.527 &\cellcolor[HTML]{efefef}- \\
\cellcolor[HTML]{efefef}FT-S &\cellcolor[HTML]{efefef}0.497 &\cellcolor[HTML]{efefef}\textbf{0.635} &\cellcolor[HTML]{efefef}0.631 &\cellcolor[HTML]{efefef}0.493 &\cellcolor[HTML]{efefef}- &\cellcolor[HTML]{efefef}0.494 &\cellcolor[HTML]{efefef}0.508 &\cellcolor[HTML]{efefef}\textbf{0.572} &\cellcolor[HTML]{efefef}0.527 &\cellcolor[HTML]{efefef}- \\
\cellcolor[HTML]{efefef}FT-M &\cellcolor[HTML]{efefef}0.543 &\cellcolor[HTML]{efefef}0.685 &\cellcolor[HTML]{efefef}\textbf{0.693} &\cellcolor[HTML]{efefef}0.493 &\cellcolor[HTML]{efefef}- &\cellcolor[HTML]{efefef}0.577 &\cellcolor[HTML]{efefef}0.573 &\cellcolor[HTML]{efefef}\textbf{0.718} &\cellcolor[HTML]{efefef}0.573 &\cellcolor[HTML]{efefef}- \\
\cellcolor[HTML]{efefef}FT-L &\cellcolor[HTML]{efefef}0.586 &\cellcolor[HTML]{efefef}\textbf{0.707} &\cellcolor[HTML]{efefef}0.696 &\cellcolor[HTML]{efefef}0.522 &\cellcolor[HTML]{efefef}- &\cellcolor[HTML]{efefef}0.640 &\cellcolor[HTML]{efefef}0.644 &\cellcolor[HTML]{efefef}\textbf{0.736} &\cellcolor[HTML]{efefef}0.692 &\cellcolor[HTML]{efefef}- \\
\cellcolor[HTML]{efefef}FT-F &\cellcolor[HTML]{efefef}0.654 &\cellcolor[HTML]{efefef}0.691 &\cellcolor[HTML]{efefef}\textbf{0.769} &\cellcolor[HTML]{efefef}0.572 &\cellcolor[HTML]{efefef}- &\cellcolor[HTML]{efefef}0.691 &\cellcolor[HTML]{efefef}0.636 &\cellcolor[HTML]{efefef}\textbf{0.734} &\cellcolor[HTML]{efefef}0.724 &\cellcolor[HTML]{efefef}- \\
\textbf{EEG} &\multicolumn{5}{c|}{SHHS (source)} &\multicolumn{5}{c}{ISRUC (target)} \\
LE &\textbf{0.581} &0.527 &0.537 &0.446 &- &\textbf{0.564} &0.479 &0.525 &0.526 &- \\
FT-S &0.551 &0.628 &0.542 &\textbf{0.681} &- &0.462 &0.496 &\textbf{0.529} &0.506 &- \\
FT-M &0.553 &0.673 &0.553 &\textbf{0.729} &- &0.498 &0.479 &0.589 &\textbf{0.580} &- \\
FT-L &0.625 &0.682 &0.603 &\textbf{0.756} &- &0.498 &0.510 &0.591 &\textbf{0.680} &- \\
FT-F &0.675 &0.690 &0.640 &\textbf{0.758} &- &0.552 &0.525 &0.583 &\textbf{0.673} &- \\
\cellcolor[HTML]{efefef}\textbf{CXR} &\multicolumn{5}{c|}{\cellcolor[HTML]{efefef}MIMIC (source)} &\multicolumn{5}{c}{\cellcolor[HTML]{efefef}CheXpert (target)} \\
\cellcolor[HTML]{efefef}LE &\cellcolor[HTML]{efefef}0.685 &\cellcolor[HTML]{efefef}0.676 &\cellcolor[HTML]{efefef}\textbf{0.732} &\cellcolor[HTML]{efefef}0.685 &\cellcolor[HTML]{efefef}0.669 &\cellcolor[HTML]{efefef}0.723 &\cellcolor[HTML]{efefef}0.745 &\cellcolor[HTML]{efefef}0.760 &\cellcolor[HTML]{efefef}0.756 &\cellcolor[HTML]{efefef}\textbf{0.812} \\
\cellcolor[HTML]{efefef}FT-S &\cellcolor[HTML]{efefef}0.685 &\cellcolor[HTML]{efefef}0.677 &\cellcolor[HTML]{efefef}\textbf{0.731} &\cellcolor[HTML]{efefef}0.685 &\cellcolor[HTML]{efefef}0.730 &\cellcolor[HTML]{efefef}0.720 &\cellcolor[HTML]{efefef}0.740 &\cellcolor[HTML]{efefef}0.732 &\cellcolor[HTML]{efefef}0.720 &\cellcolor[HTML]{efefef}\textbf{0.762} \\
\cellcolor[HTML]{efefef}FT-M &\cellcolor[HTML]{efefef}0.724 &\cellcolor[HTML]{efefef}0.724 &\cellcolor[HTML]{efefef}0.776 &\cellcolor[HTML]{efefef}0.740 &\cellcolor[HTML]{efefef}\textbf{0.787} &\cellcolor[HTML]{efefef}0.720 &\cellcolor[HTML]{efefef}0.745 &\cellcolor[HTML]{efefef}0.745 &\cellcolor[HTML]{efefef}0.720 &\cellcolor[HTML]{efefef}\textbf{0.800} \\
\cellcolor[HTML]{efefef}FT-L &\cellcolor[HTML]{efefef}0.740 &\cellcolor[HTML]{efefef}0.730 &\cellcolor[HTML]{efefef}0.782 &\cellcolor[HTML]{efefef}0.742 &\cellcolor[HTML]{efefef}\textbf{0.790} &\cellcolor[HTML]{efefef}0.720 &\cellcolor[HTML]{efefef}0.746 &\cellcolor[HTML]{efefef}0.770 &\cellcolor[HTML]{efefef}0.723 &\cellcolor[HTML]{efefef}\textbf{0.812} \\
\cellcolor[HTML]{efefef}FT-F &\cellcolor[HTML]{efefef}0.740 &\cellcolor[HTML]{efefef}0.744 &\cellcolor[HTML]{efefef}0.783 &\cellcolor[HTML]{efefef}0.789 &\cellcolor[HTML]{efefef}\textbf{0.792} &\cellcolor[HTML]{efefef}0.723 &\cellcolor[HTML]{efefef}0.746 &\cellcolor[HTML]{efefef}0.770 &\cellcolor[HTML]{efefef}0.772 &\cellcolor[HTML]{efefef}\textbf{0.813} \\
\cellcolor[HTML]{efefef}\textbf{CXR} &\multicolumn{5}{c|}{\cellcolor[HTML]{efefef}VINDR-CXR (target)} &\cellcolor[HTML]{efefef} &\cellcolor[HTML]{efefef} &\cellcolor[HTML]{efefef} &\cellcolor[HTML]{efefef} &\cellcolor[HTML]{efefef}\textbf{} \\
\cellcolor[HTML]{efefef}LE &\cellcolor[HTML]{efefef}0.546 &\cellcolor[HTML]{efefef}0.502 &\cellcolor[HTML]{efefef}0.576 &\cellcolor[HTML]{efefef}0.567 &\cellcolor[HTML]{efefef}\textbf{0.628} &\cellcolor[HTML]{efefef} &\cellcolor[HTML]{efefef} &\cellcolor[HTML]{efefef} &\cellcolor[HTML]{efefef} &\cellcolor[HTML]{efefef}\textbf{} \\
\cellcolor[HTML]{efefef}FT-S &\cellcolor[HTML]{efefef}0.533 &\cellcolor[HTML]{efefef}0.520 &\cellcolor[HTML]{efefef}0.551 &\cellcolor[HTML]{efefef}0.575 &\cellcolor[HTML]{efefef}\textbf{0.594} &\cellcolor[HTML]{efefef} &\cellcolor[HTML]{efefef} &\cellcolor[HTML]{efefef} &\cellcolor[HTML]{efefef} &\cellcolor[HTML]{efefef}\textbf{} \\
\cellcolor[HTML]{efefef}FT-M &\cellcolor[HTML]{efefef}0.561 &\cellcolor[HTML]{efefef}0.530 &\cellcolor[HTML]{efefef}0.534 &\cellcolor[HTML]{efefef}0.576 &\cellcolor[HTML]{efefef}\textbf{0.628} &\cellcolor[HTML]{efefef} &\cellcolor[HTML]{efefef} &\cellcolor[HTML]{efefef} &\cellcolor[HTML]{efefef} &\cellcolor[HTML]{efefef}\textbf{} \\
\cellcolor[HTML]{efefef}FT-L &\cellcolor[HTML]{efefef}0.562 &\cellcolor[HTML]{efefef}0.534 &\cellcolor[HTML]{efefef}0.576 &\cellcolor[HTML]{efefef}0.576 &\cellcolor[HTML]{efefef}\textbf{0.632} &\cellcolor[HTML]{efefef} &\cellcolor[HTML]{efefef} &\cellcolor[HTML]{efefef} &\cellcolor[HTML]{efefef} &\cellcolor[HTML]{efefef}\textbf{} \\
\cellcolor[HTML]{efefef}FT-F &\cellcolor[HTML]{efefef}0.576 &\cellcolor[HTML]{efefef}0.576 &\cellcolor[HTML]{efefef}\textbf{0.632} &\cellcolor[HTML]{efefef}0.579 &\cellcolor[HTML]{efefef}0.628 &\cellcolor[HTML]{efefef} &\cellcolor[HTML]{efefef} &\cellcolor[HTML]{efefef} &\cellcolor[HTML]{efefef} &\cellcolor[HTML]{efefef}\textbf{} \\
\textbf{Mammo} &\multicolumn{5}{c|}{Vindr-Mammo (source)} &\multicolumn{5}{c}{CBIS-DDSM (target)} \\
LE &0.558 &0.507 &0.500 &0.561 &\textbf{0.606} &0.464 &0.531 &0.499 &0.465 &\textbf{0.579} \\
FT-S &\textbf{0.570} &0.541 &0.509 &0.564 &0.544 &0.444 &0.513 &0.528 &0.490 &\textbf{0.541} \\
FT-M &0.587 &0.552 &0.546 &0.565 &\textbf{0.597} &0.541 &0.490 &0.502 &0.506 &\textbf{0.546} \\
FT-L &\textbf{0.601} &0.579 &0.531 &0.590 &0.594 &0.565 &0.470 &0.520 &\textbf{0.597} &0.520 \\
FT-F &0.611 &0.588 &0.542 &0.625 &\textbf{0.645} &0.487 &0.482 &0.510 &0.565 &\textbf{0.607} \\
\cellcolor[HTML]{efefef}\textbf{Derm} &\multicolumn{5}{c|}{\cellcolor[HTML]{efefef}BCN 20000 (source)} &\multicolumn{5}{c}{\cellcolor[HTML]{efefef}HAM 10000 (target)} \\
\cellcolor[HTML]{efefef}LE &\cellcolor[HTML]{efefef}0.638 &\cellcolor[HTML]{efefef}0.673 &\cellcolor[HTML]{efefef}0.605 &\cellcolor[HTML]{efefef}0.794 &\cellcolor[HTML]{efefef}\textbf{0.847} &\cellcolor[HTML]{efefef}0.609 &\cellcolor[HTML]{efefef}0.657 &\cellcolor[HTML]{efefef}0.750 &\cellcolor[HTML]{efefef}0.755 &\cellcolor[HTML]{efefef}\textbf{0.846} \\
\cellcolor[HTML]{efefef}FT-S &\cellcolor[HTML]{efefef}0.638 &\cellcolor[HTML]{efefef}0.798 &\cellcolor[HTML]{efefef}0.735 &\cellcolor[HTML]{efefef}0.666 &\cellcolor[HTML]{efefef}\textbf{0.812} &\cellcolor[HTML]{efefef}0.609 &\cellcolor[HTML]{efefef}\textbf{0.814} &\cellcolor[HTML]{efefef}0.806 &\cellcolor[HTML]{efefef}0.666 &\cellcolor[HTML]{efefef}0.809 \\
\cellcolor[HTML]{efefef}FT-M &\cellcolor[HTML]{efefef}\textbf{0.997} &\cellcolor[HTML]{efefef}0.932 &\cellcolor[HTML]{efefef}0.794 &\cellcolor[HTML]{efefef}0.777 &\cellcolor[HTML]{efefef}0.818 &\cellcolor[HTML]{efefef}0.826 &\cellcolor[HTML]{efefef}0.790 &\cellcolor[HTML]{efefef}0.760 &\cellcolor[HTML]{efefef}0.732 &\cellcolor[HTML]{efefef}\textbf{0.823} \\
\cellcolor[HTML]{efefef}FT-L &\cellcolor[HTML]{efefef}\textbf{0.998} &\cellcolor[HTML]{efefef}0.939 &\cellcolor[HTML]{efefef}0.803 &\cellcolor[HTML]{efefef}0.853 &\cellcolor[HTML]{efefef}0.836 &\cellcolor[HTML]{efefef}0.827 &\cellcolor[HTML]{efefef}0.805 &\cellcolor[HTML]{efefef}\textbf{0.904} &\cellcolor[HTML]{efefef}0.820 &\cellcolor[HTML]{efefef}0.825 \\
\cellcolor[HTML]{efefef}FT-F &\cellcolor[HTML]{efefef}\textbf{0.998} &\cellcolor[HTML]{efefef}0.955 &\cellcolor[HTML]{efefef}0.982 &\cellcolor[HTML]{efefef}0.987 &\cellcolor[HTML]{efefef}0.996 &\cellcolor[HTML]{efefef}0.827 &\cellcolor[HTML]{efefef}0.966 &\cellcolor[HTML]{efefef}0.967 &\cellcolor[HTML]{efefef}\textbf{0.986} &\cellcolor[HTML]{efefef}0.877 \\
\cellcolor[HTML]{efefef}\textbf{Derm} &\multicolumn{5}{c|}{\cellcolor[HTML]{efefef}PAD-UFES-20 (target)} &\cellcolor[HTML]{efefef} &\cellcolor[HTML]{efefef} &\cellcolor[HTML]{efefef} &\cellcolor[HTML]{efefef}\textbf{} &\cellcolor[HTML]{efefef} \\
\cellcolor[HTML]{efefef}LE &\cellcolor[HTML]{efefef}0.487 &\cellcolor[HTML]{efefef}0.589 &\cellcolor[HTML]{efefef}0.640 &\cellcolor[HTML]{efefef}0.642 &\cellcolor[HTML]{efefef}\textbf{0.648} &\cellcolor[HTML]{efefef} &\cellcolor[HTML]{efefef} &\cellcolor[HTML]{efefef} &\cellcolor[HTML]{efefef}\textbf{} &\cellcolor[HTML]{efefef} \\
\cellcolor[HTML]{efefef}FT-S &\cellcolor[HTML]{efefef}0.487 &\cellcolor[HTML]{efefef}0.600 &\cellcolor[HTML]{efefef}0.588 &\cellcolor[HTML]{efefef}0.552 &\cellcolor[HTML]{efefef}\textbf{0.629} &\cellcolor[HTML]{efefef} &\cellcolor[HTML]{efefef} &\cellcolor[HTML]{efefef} &\cellcolor[HTML]{efefef}\textbf{} &\cellcolor[HTML]{efefef} \\
\cellcolor[HTML]{efefef}FT-M &\cellcolor[HTML]{efefef}0.585 &\cellcolor[HTML]{efefef}0.591 &\cellcolor[HTML]{efefef}0.684 &\cellcolor[HTML]{efefef}\textbf{0.752} &\cellcolor[HTML]{efefef}0.647 &\cellcolor[HTML]{efefef} &\cellcolor[HTML]{efefef} &\cellcolor[HTML]{efefef} &\cellcolor[HTML]{efefef}\textbf{} &\cellcolor[HTML]{efefef} \\
\cellcolor[HTML]{efefef}FT-L &\cellcolor[HTML]{efefef}0.585 &\cellcolor[HTML]{efefef}0.594 &\cellcolor[HTML]{efefef}0.658 &\cellcolor[HTML]{efefef}\textbf{0.758} &\cellcolor[HTML]{efefef}0.650 &\cellcolor[HTML]{efefef} &\cellcolor[HTML]{efefef} &\cellcolor[HTML]{efefef} &\cellcolor[HTML]{efefef}\textbf{} &\cellcolor[HTML]{efefef} \\
\cellcolor[HTML]{efefef}FT-F &\cellcolor[HTML]{efefef}0.596 &\cellcolor[HTML]{efefef}0.598 &\cellcolor[HTML]{efefef}0.656 &\cellcolor[HTML]{efefef}\textbf{0.794} &\cellcolor[HTML]{efefef}0.660 &\cellcolor[HTML]{efefef} &\cellcolor[HTML]{efefef} &\cellcolor[HTML]{efefef} &\cellcolor[HTML]{efefef}\textbf{} &\cellcolor[HTML]{efefef} \\
\textbf{Fundus} &\multicolumn{5}{c|}{Messidor-2 (source)} &\multicolumn{5}{c}{APTOS 2019 (target)} \\
LE &0.677 &0.660 &0.451 &\textbf{0.741} &0.741 &0.474 &0.496 &0.496 &0.440 &\textbf{0.641} \\
FT-S &0.647 &0.625 &0.441 &\textbf{0.713} &0.679 &0.453 &0.422 &0.552 &0.381 &\textbf{0.602} \\
FT-M &0.731 &0.739 &0.481 &0.752 &\textbf{0.800} &0.514 &0.411 &0.539 &0.420 &\textbf{0.593} \\
FT-L &0.831 &0.870 &0.507 &\textbf{0.891} &0.890 &0.535 &0.450 &0.573 &0.449 &\textbf{0.683} \\
FT-F &0.983 &1.000 &0.557 &\textbf{1.000} &\textbf{1.000} &0.472 &0.417 &0.564 &0.476 &\textbf{0.673} \\
\textbf{Fundus} &\multicolumn{5}{c|}{Jinchi Medical University (target)} & & & & & \\
LE &0.523 &0.532 &0.500 &0.546 &\textbf{0.587} & & & & & \\
FT-S &0.583 &\textbf{0.612} &0.500 &0.495 &0.505 & & & & & \\
FT-M &\textbf{0.556} &0.537 &0.500 &0.533 &0.546 & & & & & \\
FT-L &0.561 &0.653 &0.488 &0.546 &\textbf{0.679} & & & & & \\
FT-F &0.570 &\textbf{0.632} &0.499 &0.529 &0.602 & & & & & \\
\cellcolor[HTML]{efefef}\textbf{LDCT} &\multicolumn{5}{c|}{\cellcolor[HTML]{efefef}LIDC-IDRI (source)} &\multicolumn{5}{c}{\cellcolor[HTML]{efefef}LNDb (target)} \\
\cellcolor[HTML]{efefef}LE &\cellcolor[HTML]{efefef}0.607 &\cellcolor[HTML]{efefef}\textbf{0.783} &\cellcolor[HTML]{efefef}0.779 &\cellcolor[HTML]{efefef}0.760 &\cellcolor[HTML]{efefef}- &\cellcolor[HTML]{efefef}0.578 &\cellcolor[HTML]{efefef}0.495 &\cellcolor[HTML]{efefef}\textbf{0.621} &\cellcolor[HTML]{efefef}0.615 &\cellcolor[HTML]{efefef}- \\
\cellcolor[HTML]{efefef}FT-S &\cellcolor[HTML]{efefef}0.599 &\cellcolor[HTML]{efefef}\textbf{0.646} &\cellcolor[HTML]{efefef}0.637 &\cellcolor[HTML]{efefef}0.578 &\cellcolor[HTML]{efefef}- &\cellcolor[HTML]{efefef}\textbf{0.576} &\cellcolor[HTML]{efefef}0.515 &\cellcolor[HTML]{efefef}0.574 &\cellcolor[HTML]{efefef}0.524 &\cellcolor[HTML]{efefef}- \\
\cellcolor[HTML]{efefef}FT-M &\cellcolor[HTML]{efefef}0.617 &\cellcolor[HTML]{efefef}0.697 &\cellcolor[HTML]{efefef}0.717 &\cellcolor[HTML]{efefef}\textbf{0.743} &\cellcolor[HTML]{efefef}- &\cellcolor[HTML]{efefef}0.622 &\cellcolor[HTML]{efefef}0.514 &\cellcolor[HTML]{efefef}0.625 &\cellcolor[HTML]{efefef}\textbf{0.660} &\cellcolor[HTML]{efefef}- \\
\cellcolor[HTML]{efefef}FT-L &\cellcolor[HTML]{efefef}0.791 &\cellcolor[HTML]{efefef}\textbf{0.825} &\cellcolor[HTML]{efefef}0.812 &\cellcolor[HTML]{efefef}0.810 &\cellcolor[HTML]{efefef}- &\cellcolor[HTML]{efefef}0.638 &\cellcolor[HTML]{efefef}0.620 &\cellcolor[HTML]{efefef}\textbf{0.649} &\cellcolor[HTML]{efefef}0.647 &\cellcolor[HTML]{efefef}- \\
\cellcolor[HTML]{efefef}FT-F &\cellcolor[HTML]{efefef}0.797 &\cellcolor[HTML]{efefef}\textbf{0.821} &\cellcolor[HTML]{efefef}0.817 &\cellcolor[HTML]{efefef}0.818 &\cellcolor[HTML]{efefef}- &\cellcolor[HTML]{efefef}0.623 &\cellcolor[HTML]{efefef}0.597 &\cellcolor[HTML]{efefef}\textbf{0.661} &\cellcolor[HTML]{efefef}0.647 &\cellcolor[HTML]{efefef}- \\
\bottomrule
\end{tabular}
\end{table*}

%Please add the following packages if necessary:
%\usepackage{booktabs, multirow} % for borders and merged ranges
%\usepackage{soul}% for underlines
%\usepackage[table]{xcolor} % for cell colors
%\usepackage{changepage,threeparttable} % for wide tables
%If the table is too wide, replace \begin{table}[!htp]...\end{table} with
%\begin{adjustwidth}{-2.5 cm}{-2.5 cm}\centering\begin{threeparttable}[!htb]...\end{threeparttable}\end{adjustwidth}
\begin{table*}[!htp]\centering
\caption{AUROC achieved for the 2D mammogram and fundus images modalities, with additional results reported for performing MAE pre-training (the best performing SSL algorithm on source data in both modalities). For the IN+MAE results columns, we also performed additional hyperparameter tuning, trying learning rates of 1e-3, 1e-4, and 1e-5 combinations across the MAE pre-training and transfer learning stages. For each dataset and transfer learning setup, we bold the max AUROC achieved across the different pre-training methods, with the exception of the result for Jinchi Medical University in the finetune-small transfer setting, where e-Mix still outperforms IN+MAE.}\label{tab:res-inpt}
\scriptsize
\begin{tabular}{lcccc|cccc|ccccc}\toprule
&Scratch &MAE &IN+MAE &IN &Scratch &MAE &IN+MAE &IN &Scratch &MAE &IN+MAE &IN \\
\cellcolor[HTML]{efefef}\textbf{Mammo} &\multicolumn{4}{c|}{\cellcolor[HTML]{efefef}Vindr-Mammo (source)} &\multicolumn{4}{c|}{\cellcolor[HTML]{efefef}CBIS-DDSM (target)} &\cellcolor[HTML]{efefef} &\cellcolor[HTML]{efefef} &\cellcolor[HTML]{efefef} &\cellcolor[HTML]{efefef} \\
\cellcolor[HTML]{efefef}LE &\cellcolor[HTML]{efefef}0.558 &\cellcolor[HTML]{efefef}0.561 &\cellcolor[HTML]{efefef}\textbf{0.610} &\cellcolor[HTML]{efefef}0.606 &\cellcolor[HTML]{efefef}0.464 &\cellcolor[HTML]{efefef}0.465 &\cellcolor[HTML]{efefef}\textbf{0.584} &\cellcolor[HTML]{efefef}0.579 &\cellcolor[HTML]{efefef} &\cellcolor[HTML]{efefef} &\cellcolor[HTML]{efefef} &\cellcolor[HTML]{efefef} \\
\cellcolor[HTML]{efefef}FT-S &\cellcolor[HTML]{efefef}\textbf{0.570} &\cellcolor[HTML]{efefef}0.564 &\cellcolor[HTML]{efefef}\textbf{0.575} &\cellcolor[HTML]{efefef}0.544 &\cellcolor[HTML]{efefef}0.444 &\cellcolor[HTML]{efefef}0.490 &\cellcolor[HTML]{efefef}0.516 &\cellcolor[HTML]{efefef}\textbf{0.541} &\cellcolor[HTML]{efefef} &\cellcolor[HTML]{efefef} &\cellcolor[HTML]{efefef} &\cellcolor[HTML]{efefef} \\
\cellcolor[HTML]{efefef}FT-M &\cellcolor[HTML]{efefef}0.587 &\cellcolor[HTML]{efefef}0.565 &\cellcolor[HTML]{efefef}\textbf{0.605} &\cellcolor[HTML]{efefef}0.597 &\cellcolor[HTML]{efefef}0.541 &\cellcolor[HTML]{efefef}0.506 &\cellcolor[HTML]{efefef}\textbf{0.592} &\cellcolor[HTML]{efefef}0.546 &\cellcolor[HTML]{efefef} &\cellcolor[HTML]{efefef} &\cellcolor[HTML]{efefef} &\cellcolor[HTML]{efefef} \\
\cellcolor[HTML]{efefef}FT-L &\cellcolor[HTML]{efefef}\textbf{0.601} &\cellcolor[HTML]{efefef}0.590 &\cellcolor[HTML]{efefef}0.605 &\cellcolor[HTML]{efefef}0.594 &\cellcolor[HTML]{efefef}0.565 &\cellcolor[HTML]{efefef}\textbf{0.597} &\cellcolor[HTML]{efefef}0.577 &\cellcolor[HTML]{efefef}0.520 &\cellcolor[HTML]{efefef} &\cellcolor[HTML]{efefef} &\cellcolor[HTML]{efefef} &\cellcolor[HTML]{efefef} \\
\cellcolor[HTML]{efefef}FT-F &\cellcolor[HTML]{efefef}0.611 &\cellcolor[HTML]{efefef}0.625 &\cellcolor[HTML]{efefef}0.616 &\cellcolor[HTML]{efefef}\textbf{0.645} &\cellcolor[HTML]{efefef}0.487 &\cellcolor[HTML]{efefef}0.565 &\cellcolor[HTML]{efefef}0.492 &\cellcolor[HTML]{efefef}\textbf{0.607} &\cellcolor[HTML]{efefef} &\cellcolor[HTML]{efefef} &\cellcolor[HTML]{efefef} &\cellcolor[HTML]{efefef} \\
\textbf{Fundus} &\multicolumn{4}{c|}{Messidor-2 (source)} &\multicolumn{4}{c|}{APTOS 2019 (target)} &\multicolumn{4}{c}{Jinchi Medical University (target)} \\
LE &0.677 &0.741 &\textbf{0.751} &0.741 &0.474 &0.440 &0.553 &\textbf{0.641} &0.523 &0.546 &0.562 &\textbf{0.587} \\
FT-S &0.647 &\textbf{0.713} &0.663 &0.679 &0.453 &0.381 &0.429 &\textbf{0.602} &0.583 &0.495 &\textbf{0.542} &0.505 \\
FT-M &0.731 &0.752 &0.782 &\textbf{0.800} &0.514 &0.420 &0.477 &\textbf{0.593} &0.556 &0.533 &\textbf{0.675} &0.546 \\
FT-L &0.831 &\textbf{0.891} &0.882 &0.890 &0.535 &0.449 &0.608 &\textbf{0.683} &0.561 &0.546 &0.661 &\textbf{0.679} \\
FT-F &0.983 &\textbf{1.000} &\textbf{1.000} &\textbf{1.000} &0.472 &0.476 &0.621 &\textbf{0.673} &0.570 &0.529 &\textbf{0.674} &0.602 \\
\bottomrule
\end{tabular}
\end{table*}

\section{Methods Details}

Our domain-agnostic transformer architecture contains 9.7M parameters, the same as the ImageNet-pretrained ViT-T we use to compare performance for 2D modalities. The domain-agnostic architecture is the same as that used in \cite{Tamkin2021-jw} and all architecture implementations can be found in our repository here: {\small\url{https://github.com/rajpurkarlab/BenchMD/tree/main/src/models}}. We trained in single-gpu mode with A10G Tensor Core GPUs, and the batch size used in all experiments was 64, with the exception of LDCT experiments, where we used a batch size of 32. Each pretraining run takes roughly 30 hours, and each transfer learning run takes up to 1 day, with most experiments finishing in under 8 hours.

\section{Dataset Input and Label Preprocessing}
\label{sec:preprocess}

\subsection{ECG}

\paragraph{Preprocessing} All data was exhaustively cropped to 10 second segments, sampled at 500 Hz, making each input a 1D vector of 2500 components with 12 channels. Leftover segments of less that 10 seconds were dropped. 
We made significant re-labeling progress to create an unified machine learning task for each of the datasets. Our finally 7 classes include: Normal, CD (Conduction Disturbance), HYP (Hypertrophy), MI (Myocardial Infarction), STTC (ST-T wave Change-ischemia), A. Fib/ Aflutter (Atrial fibrillation/ Atrial flutter), and Other. Below is a name mapping from each original dataset to our new label formulation.

\paragraph{Task Standardization} 

\begin{table*}[!htp]\centering
\caption{PTB-XL Label Mappings}\label{tab:ptbxllabels}
\scriptsize
\begin{tabular}{p{1.5cm}p{11cm}}\toprule
Class Name & PTB-XL Labels Included \\
\hline
Normal & NORM, SARRH, SBRAD, SR, STACH \\
CD & AVB, 1AVB, 2AVB, 3AVB, CD, CLBBB, CRBBB, ILBBB, IRBBB, IVCB IVCD, LAFB, LAFB/LPFB, LPFB, LPR, PSVT, SVARR, SVTAC, WPW \\
HYP	& HYP, ALAD, LAD, LAO/LAE, LVH, RAD, RHV, RVH, RAO/RAE, SEHYP, VCLVH \\
MI	& AMI, ALMI,ASMI, ILMI, IMI, INJAL, INJIL, INJLA, INVT, IPLMI, IPMI, LMI, MI, PMI \\
STTC & ANEUR, DIG, EL, ISC\_, ISCA, ISCAL, ISCAN, ISCAS, ISCI, ISCIL, ISCIN, ISCLA, LNGQT, NDT, NST\_, NT\_, STD\_, STE\_, STTC, TAB\_ \\
A. Fib/ Aflutter & AFIB, AFLT \\
Other & ABQRS, ARAD, AXL, AXR, BIGU, HVOLT, LOWT, LVOLT, PACE, \\
& PAC, PRC(S), PVC, QWAVE, SAG, and TRIGU \\
\bottomrule
\end{tabular}
\end{table*}

\begin{table*}[!htp]\centering
\caption{Chapman-Shaoxing Label Mappings}\label{tab:chapmanshaoxinglabels}
\scriptsize
\begin{tabular}{p{1.5cm}p{11cm}}\toprule
Class Name & Chapman-Shaoxing Labels Included \\
\hline
Normal & NORM, SB, SR, ST \\
CD & 1AVB, 2AVB2, AVB, AVNRT, AT, CAVB, CLBBB, IIAVBI, IVB, JEB, JPT, Nonspecific BBB, PRIE, PRWP, PWC, SAAWR, SVT, VEB, VET, VPB, VPE, WAVN, WPW \\
HYP & ALS, ARS, CR, LVH, LVHV, RAH, RAVC, RVH \\
MI & MILW \\
STTC & STDD, STE, STTC, STTU, TTW, TWO \\
A. Fib/ Aflutter & AF, AFIB \\
Other & ABI, APB, AQW, ERV, FQRS, LVQRSCL, LVQRSLL, PTW, UW, VB \\
\bottomrule
\end{tabular}
\end{table*}

\begin{table*}[!htp]\centering
\caption{Georgia ECG Label Mappings}\label{tab:georgialabels}
\scriptsize
\begin{tabular}{p{1.5cm}p{11cm}}\toprule
Class Name & Georgia ECG Labels Included \\
\hline
Normal & Bradycardia, sinus arrhythmia, sinus bradycardia, sinus rhythm, sinus tachycardia\\
CD& 1st degree av block, 2nd degree av block, accelerated idioventricular rhythm, accelerated junctional rhythm, Atrial pacing pattern, Atrial tachycardia, AV block, Brady Tachy syndrome, Bundle branch block, Cardiac dysrhythmia, complete heart block, complete right bundle branch block, congenital incomplete atrioventricular heart block, diffuse intraventricular block, ectopic rhythm, idioventricular rhythm, incomplete left bundle branch block, incomplete right bundle branch block, junctional escape, junctional premature complex, junctional tachycardia,left anterior fascicular block, left bundle branch block, left posterior fascicular block, mobitz type 2 second degree atrioventricular block, mobitz type i wenckebach atrioventricular block, multifocal atrial tachycardia, paroxysmal supraventricular tachycardia, paroxysmal ventricular tachycardia, partial atrioventricular block 2:1, prolonged pr interval,right bundle branch block, shortened pr interval,sinus node dysfunction, supraventricular bigeminy, supraventricular premature beats, supraventricular tachycardia, ventricular ectopic beats, ventricular escape beat, ventricular escape rhythm, ventricular fibrillation, ventricular flutter, ventricular pacing pattern, ventricular preexcitation, ventricular tachycardia, ventricular trigeminy, wandering atrial pacemaker, wolff parkinson white pattern
HYP	trial hypertrophy, left atrial abnormality, left atrial enlargement, left atrial hypertrophy, left axis deviation, left ventricular hypertrophy, left ventricular strain, r wave abnormal, right atrial abnormality, right atrial hypertrophy, right axis deviation, right ventricular hypertrophy, ventricular hypertrophy \\
MI & Acute myocardial infarction, Acute myocardial ischemia, Anterior ischemia, chronic myocardial ischemia, inferior ischaemia, inferior st segment depression, lateral ischaemia, myocardial infarction, myocardial ischemia, old myocardial infarction \\
STTC & coronary heart disease, electrical alternans, high t voltage, nonspecific st t abnormality, s t changes, st depression, st elevation, st interval abnormal, t wave abnormal, t wave inversion \\
A. Fib/ Aflutter & Atrial fibrillation, Atrial fibrillation and flutter, Atrial flutter, chronic atrial fibrillation, paroxysmal atrial fibrillation, rapid atrial fibrillation\\
Other & Abnormal QRS, Atrial bigeminy, Blocked premature atrial contraction, Brugada syndrome, chronic rheumatic pericarditis, decreased qt interval, early repolarization, ecg artefacts, fusion beats, heart failure, indeterminate cardiac axis, isorhythmic dissociation, low qrs voltages, low qrs voltages in the limb leads, low qrs voltages in the precordial leads, non-specific interatrial conduction block, nonspecific intraventricular conduction disorder, pacing rhythm, paired ventricular premature complexes, premature atrial contraction, premature ventricular complexes, premature ventricular contractions, prolonged qt interval, qwave abnormal, suspect arm ecg leads reversed, tall u wave, transient ischemic attack, u wave abnormal, ventricular bigeminy\\
\bottomrule
\end{tabular}
\end{table*}

\begin{table*}[!htp]\centering
\caption{CPSC Label Mappings}\label{tab:cpsclabels}
\scriptsize
\begin{tabular}{p{1.5cm}p{11cm}}\toprule
Class Name & CPSC Labels Included \\
\hline
Normal & sinus rhythm \\
CD & 1st degree av block, atrial fibrillation, right bundle branch block, ventricular ectopics \\
HYP & hypertrophy\\
MI & MI \\
STTC & st depression, st elevation\\
A. Fib/ Aflutter & AF, AFIB \\
Other & premature atrial contraction \\
\bottomrule
\end{tabular}
\end{table*}

The label consolidation for each ECG dataset under the the 7-class task is given in Tables \ref{tab:ptbxllabels}-\ref{tab:cpsclabels}. The distribution of classes for ECG datasets is shown in Table \ref{tab:ecg}.

\begin{table*}[!htp]\centering
\caption{Class distributions for ECG datasets.}\label{tab:ecg}
\scriptsize
\begin{tabular}{lrrrrrr}\toprule
&\multicolumn{2}{c}{PTB-XL (source)} &Chapman-Shaoxing (target) &Georgia (target) &CPSC (target) \\
Class &Training Split &Validation Split &Validation Split &Validation Split &Validation Split \\
Normal &9222 (52.77\%) &2322 (53.24\%) &1129 (55.05\%) &725 (35.07\%) &190 (13.8\%) \\
Conduction Disturbance &1386 (7.93\%) &348 (7.98\%) &249 (12.14\%) &240 (11.61\%) &717 (52.07\%) \\
Myocardial Infarction &1285 (7.35\%) &333 (7.64\%) &2 (0.098\%) &82 (3.97\%) &5 (0.36\%) \\
Ischemic ST-T Changes &1661 (9.5\%) &420 (9.63\%) &260 (12.68\%) &437 (21.14\%) &213 (15.47\%) \\
Other &1462 (8.37\%) &360 (8.25\%) &33 (1.61\%) &263 (12.72\%) &116 (8.42\%) \\
Atrial fibrillation/atrial flutter &475 (2.72\%) &103 (2.36\%) &232 (11.31\%) &2 (0.097\%) &131 (9.51\%) \\
Hypertrophy &1985 (11.36\%) &475 (10.89\%) &146 (7.12\%) &318 (15.38\%) &5 (0.36\%) \\
Total \# Examples &17476 &4361 &2051 &2067 &1377 \\
\bottomrule
\end{tabular}
\end{table*}

\subsection{EEG}

\paragraph{Preprocessing} The Sleep Heart Health Study dataset consists of two rounds of polysomnographic recordings (SHHS-1 and SHHS-2) sampled at 125 Hz, and we only use SHHS-1, containing 5,793 records over two channels (C4-A1 and C3-A2). Recordings are manually classified into one of six classes (W, N1, N2, N3, N4 and REM). In SHHS, we have an additional stage N4, which we merge with the N3 stage, matching the five stages of sleep according to the American Academy of Sleep Medicine (AASM) \cite{sridhar2020deep}. Each channel of the EEG recording is a vector of  3750 components, (125 Hz $\times$ 30 second recording), and one patient has multiple recording epochs of 30 seconds.

The recordings from the transfer dataset (ISRUC) consist of channels C3 and C4, which were also segmented into epochs of 30 seconds. ISRUC dataset was downsampled to 125Hz from the original 150Hz to match SHHS.

The distribution of classes for EEG datasets is shown in Table \ref{tab:eeg}.

\begin{table*}[!htp]\centering
\caption{Class distributions for EEG datasets.}\label{tab:eeg}
\scriptsize
\begin{tabular}{lrrrr}\toprule
&\multicolumn{2}{c}{SHHS (source)} &ISRUC (target) \\
Class &Training Split &Validation Split &Validation Split \\
Wake &1172690 (28.8\%) &294869 (29.04\%) &4814 (26.44\%) \\
Non-REM Stage 1 &152066 (3.74\%) &38478 (3.79\%) &2490 (13.68\%) \\
Non- REM Stage 2 &1668940 (41\%) &411170 (40.5\%) &5605 (30.78\%) \\
Non-REM Stage 3 &478497 (11.75\%) &121076 (11.92\%) &2944 (16.17\%) \\
REM &598946 (14.71\%) &149734 (14.75\%) &2175 (11.95\%) \\
Total \# Examples &4071139 &1015327 &18208 \\
\bottomrule
\end{tabular}
\end{table*}

\subsection{Chest X-Rays}

\paragraph{Preprocessing} During training, we load each 3-channel JPG image, transform it to a 1-channel grayscale image (except for the VinDr-CXR dataset, where we extract the grayscale pixel array directly from the DICOM file), resize the longer size to 224 pixels while maintaining the image’s aspect ratio, perform per-channel standardization based on the training set statistics, and pad the image with zeros to get a 224$\times$224 final grayscale image. We selected the five competition categories from CheXpert \cite{irvin2019chexpert} as our classes: Atelectasis, Cardiomegaly, Consolidation, Edema, and Pleural Effusion.

For VinDR-CXR dataset, We perform the standard pixel array extraction from the DICOM files:

\begin{enumerate}
    \item Extract the single-channel grayscale ``pixel\_array" from the DICOM file. 
    \item Scale the pixel array by a factor of ``RescaleSlope" attribute and add the value of the ``RescaleIntercept" to every pixel, if these attributes are available.
    \item Rescale the array to pixel values between 0 and 255.
    \item Invert the pixels if the ``PhotometricInterpretation" attribute is set to ``MONOCHROME1."
\end{enumerate}

The distribution of classes for chest x-ray datasets is shown in Table \ref{tab:cxr}.

\begin{table*}[!htp]\centering
\caption{Class distributions for chest x-ray datasets.}\label{tab:cxr}
\scriptsize
\begin{tabular}{lrrrrr}\toprule
&\multicolumn{2}{c}{MIMIC (source)} &CheXpert (target) &VINDR-CXR (target) \\
Class (Multi-label) &Training Split Occurrences &Validation Split Occurrences &Validation Split Occurrences &Validation Split Occurrences \\
Atelectasis &1603 (20.04\%) &425 (21.25\%) &233 (31.74\%) &86 (2.87\%) \\
Cardiomegaly &1589 (19.86\%) &445 (22.25\%) &219 (29.84\%) &309 (10.3\%) \\
Consolidation &409 (5.11\%) &108 (5.4\%) &62 (8.45\%) &96 (3.2\%) \\
Edema &925 (11.56\%) &294 (14.7\%) &23 (3.13\%) &10 (0.33\%) \\
Pleural Effusion &1930 (24.13\%) &576 (28.8\%) &171 (23.29\%) &111 (3.7\%) \\
Total \# Examples &8000 &2000 &734 &3000 \\
\bottomrule
\end{tabular}
\end{table*}

\subsection{Mammograms}

\paragraph{Preprocessing} The mammography data is distributed in the Digital Imaging and Communications in Medicine (DICOM) file format, so to improve data access speeds during training, we preprocess the data into JPG format. We first perform the standard pixel array extraction from the DICOM files:

\begin{enumerate}
    \item Extract the single-channel grayscale ``pixel\_array" from the DICOM file. 
    \item Scale the pixel array by a factor of ``RescaleSlope" attribute and add the value of the ``RescaleIntercept" to every pixel, if these attributes are available.
\end{enumerate}

From here, we save the pixel arrays as JPGs:

\begin{enumerate}
    \item Rescale the array to pixel values between 0 and 255.
    \item Invert the pixels if the ``PhotometricInterpretation" attribute is set to ``MONOCHROME1."
    \item Save the pixel array as a JPEG using the Python Imaging Library (PIL).
\end{enumerate}

During training, we load each JPG image, resize the longer size to 224 pixels while maintaining the image’s aspect ratio, zero the mean using the training set mean, and pad the image with zeros to get a 224$\times$224 final grayscale image. We discard a handful of datapoints belonging to BI-RADS 0 or BI-RADS 6, since these classes are not present in both datasets.

The distribution of classes for mammogram datasets is shown in Table \ref{tab:mammo}.

\begin{table*}[!htp]\centering
\scriptsize
\begin{tabular}{lrrrr}\toprule
&\multicolumn{2}{c}{VinDr-Mammo (source)} &CBIS-DDSM (target) \\
Class &Training Split &Validation Split &Validation Split \\
BI-RADS 1 &10724 (67.02\%) &2682 (67.05\%) &2 (0.54\%) \\
BI-RADS 2 &3742 (23.38\%) &934 (23.35\%) &15 (4.10\%) \\
BI-RADS 3 &744 (4.65\%) &186 (4.65\%) &78 (21.36\%) \\
BI-RADS 4 &610 (3.81\%) &152 (3.8\%) &188 (51.50\%) \\
BI-RADS 5 &180 (1.12\%) &46 (1.15\%) &82 (22.46\%) \\
Total \# Examples &16000 &4000 &365 \\
\bottomrule
\end{tabular}
\caption{Class distributions for mammogram datasets.}\label{tab:mammo}
\end{table*}

\subsection{Dermoscopic Images}

\paragraph{Preprocessing} During training, we load in each 3-channel JPG image, resize the longer size to 224 pixels while maintaining the image’s aspect ratio, perform per-channel standardization based on the training set statistics, and pad the image with zeros to get a 224$\times$224 final RGB image. 

\paragraph{Task Standardization} We reformulated the labels across each datasets to an unified 5 class classification task: AKIEC (includes actinic keratoses, intraepithelial carcinoma, and squamous cell carcinoma as all of these are with the continuum of squamous cell carcinoma), BCC (basal cell carcinoma), MEL (melanoma), NEV (nevus), and Other diseases ( dermatofibroma, etc). 

BCN20000 includes annotations for BCC, SCC, ACK, MEL, NEV, Dermatofibroma, Vascular lesion, and seborrheic keratosis. We grouped SCC and ACK into AKIEC, and grouped Dermatofibroma, Vascular lesion, and seborrheic keratosis into Other.

HAM10000 includes annotations for BCC, AKIEC, MEL, NV, BKL, Dermatofibroma, and VASC. We grouped BKL (benign keratosis-like lesions: solar lentigines / seborrheic keratoses), Dermatofibroma, and VASC into Other.

PAD-UFES-20 includes annotations for BCC, SCC, ACK, MEL, NEV, and Seborrheic Keratosis. We grouped SCC and ACK into AKIEC, and grouped Seborrheic Keratosis into Other. 

The distribution of classes for dermoscopic datasets is shown in Table \ref{tab:derm}.

\begin{table*}[!htp]\centering
\scriptsize
\begin{tabular}{lrrrrr}\toprule
&\multicolumn{2}{c}{BCN 20000 (source)} &HAM 10000 (target) &PAD-UFES-20 (target) \\
Class &Training Split &Validation Split &Validation Split &Validation Split \\
MEL &3618 (17.85\%) &904 (17.84\%) &223 (11.13\%) &10 (2.18\%) \\
NEV &10300 (50.83\%) &2575 (50.83\%) &1341 (66.95\%) &49 (10.68\%) \\
BCC &2658 (13.12\%) &665 (13.13\%) &103 (5.14\%) &169 (36.82\%) \\
AKIEC &1196 (5.9\%) &299 (5.9\%) &65 (3.25\%) &184 (40.09\%) \\
Other diseases &2493 (12.3\%) &623 (12.3\%) &271 (13.53\%) &47 (10.24\%) \\
Total \# Examples &20265 &5066 &2003 &459 \\
\bottomrule
\end{tabular}
\caption{Class distributions for dermoscopic image datasets.}\label{tab:derm}
\end{table*}

\subsection{Fundus Images}

\paragraph{Preprocessing} During training, we load each 3-channel JPG image, resize the longer size to 224 pixels while maintaining the image’s aspect ratio, perform per-channel standardization based on the training set statistics, and pad the image with zeros to get a 224$\times$224 final RGB image.

\paragraph{Task Standardization} For each eye fundus image, we formulate a single-label task of predicting the severity of diabetic retinopathy (DR) in the image using the International Clinic Diabetic Retinopathy (ICDR) classification scale, which classifies DR on a five-stage severity scale from 0-4. The five ratings in order of increasing severity are (0) no apparent retinopathy (NDR), (1) mild nonproliferative retinopathy (NPDR), (2) moderate NPDR, (3) severe NPDR, and (4) proliferative diabetic retinopathy (PDR). This is the scale used by the Messidor-2 and APTOS 2019 datasets. The scale can be simplified into the modified Davis scale of three stages: NDR, simple diabetic retinopathy (SDR), pre-proliferative retinopathy (PPDR), and PDR, with ICDR rating 0 corresponding to NDR, ICDR ratings 1 and 2 corresponding to SDR, ICDR rating 3 corresponding to PPDR, and ICDR rating 4 corresponding to PDR. This is the label set used by the Jinchi Medical University dataset\cite{Takahashi2017-nn}. When testing the performance of our model on this dataset, we first run prediction on the 5-class task. Then if the target label is SDR and the predicted label is either 1 or 2, then we count it as a correct prediction when computing AUROC.

The distribution of classes for fundus datasets is shown in Table \ref{tab:fundus}.

\begin{table*}[!htp]\centering
\scriptsize
\begin{tabular}{lrrrrr}\toprule
&\multicolumn{2}{c}{Messidor-2 (source)} &APTOS 2019 (target) &Jinchi (target) \\
Class &Training Split &Validation Split &Validation Split &Validation Split \\
Class 0 &813 (58.32\%) &204 (58.28\%) &361 (49.24\%) &1313 (66.01\%) \\
Class 1 &216 (15.49\%) &54 (15.42\%) &74 (10.09\%) &\multirow{2}{*}{423 (21.26\%)} \\
Class 2 &277 (19.87\%) &70 (20\%) &200 (27.28\%) & \\
Class 3 &60 (4.30\%) &15 (4.28\%) &39 (5.32\%) &92 (4.62\%) \\
Class 4 &28 (2.01\%) &7 (2\%) &59 (8.04\%) &161 (8.09\%) \\
Total \# Examples &1394 &305 &733 &1989 \\
\bottomrule
\end{tabular}
\caption{Class distributions for fundus image datasets.}\label{tab:fundus}
\end{table*}

\subsection{LDCT}

\paragraph{Preprocessing} The LIDC data is distributed in DICOM file format. We perform the following preprocessing step:

\begin{enumerate}
    \item Extract ``pixel\_array" from the DICOM file. 
    \item Scale the pixel array by a factor of ``RescaleSlope" attribute and add the value of the ``RescaleIntercept" to every pixel.
    \item Resize each pixel array to 256$\times$256.
    \item Adjust pixel array to the PE viewing window (window\_center=-600, window\_width=1500). We only keeping pixel values within a range of [window center + window width/2, window center - window width/2].
    \item Rescales the pixels into the range 0-1.
    \item Save pixels in an HDF5 file to improve I/O.
\end{enumerate}

The LNDb data is very different and stored in raw format. We perform the following preprocessing steps: 

\begin{enumerate}
    \item Adjust pixel array to the viewing window (window\_center=400, window\_width=1000). We only keep pixel values within a range of [window center + window width/2, window center - window width/2].
    \item Rescale the pixels into the range 0-1.
    \item Resize each pixel array to 256$\times$256.
    \item Map 3d segmentations to the CT scans from real world coordinates to array level coordinates and generate labels.
\end{enumerate}

During training, we load a window of 24 slices from a study and center-crop the image to 224$\times$224. 

The distribution of classes for LDCT datasets is shown in Table \ref{tab:ldct}.

\begin{table*}[!htp]\centering
\scriptsize
\begin{tabular}{lrrrr}\toprule
&\multicolumn{2}{c}{LIDC-IDRI (source)} &LNDb (target) \\
Class (Multi label) &Training Split Occurrences &Validation Split Occurrences &Validation Split Occurrences \\
Small Nodule Exists &36 (5.05\%) &6 (3.97\%) &81 (35.37\%) \\
Large Nodule Exists &346 (48.53\%) &84 (55.63\%) &203 (88.65\%) \\
Total \# Examples &713 &151 &229 \\
\bottomrule
\end{tabular}
\caption{Class distributions for LDCT datasets.}\label{tab:ldct}
\end{table*}

\end{document}